\documentclass[letterpaper, 10 pt, conference]{ieeeconf}  

\IEEEoverridecommandlockouts                              

\usepackage{amsmath} 
\usepackage{amssymb}  
\usepackage[T1]{fontenc}
\usepackage{graphicx}
\usepackage{booktabs}
\usepackage{array}

\title{\LARGE \bf
MOSAIC-SV: Real-Time Adaptive Identification of Vessel Dynamics for the Control and Deployment of Aquatic Robots
}

\author{Wensen Liu$^{1}$, Jerry Peng$^{1}$, Shravani Vedagiri$^{1}$, and Aaron M. Johnson$^{1}$
\thanks{*This material is based upon work supported by the Department of War SMART Scholarship-for-Service Program. Any opinions, findings, and conclusions or recommendations expressed in this material are those of the author(s) and do not necessarily reflect the views of the Department of War.}
\thanks{$^{1}$ Mechanical Engineering Department, Carnegie Mellon University, Pittsburgh, PA 15213, USA, {\tt\small \{wensenl, amj1\}@andrew.cmu.edu}}%
}

\begin{document}
\bstctlcite{IEEEexample:BSTcontrol} 

\maketitle
\thispagestyle{empty}
\pagestyle{empty}

\begin{abstract}
Model-based control of an aquatic robotic platform depends on a hydrodynamic model that is costly to identify and specific to the hull, payload, and conditions it was measured in. Here, we present MOSAIC-SV, a deployable real-time adaptive dynamics identification and control system that identifies a control-sufficient dynamics model from a spec-sheet engineering prior, without dedicated identification trials, and re-estimates it at every control step of a closed-loop mission while the controller plans on it. A physically admissible unscented Kalman filter re-estimates hydrodynamic, disturbance, and actuator parameters at every control step of the closed-loop mission; while a command-dependent consider projection withholds corrections the current command cannot attribute between actuator effectiveness and external force; and a model predictive path integral controller plans on the current estimate. In simulation on a CyberShip II plant, MOSAIC-SV recovers the transit performance of the calibrated model under static mismatch and transient changes, and stays within $20\%$ of its own transit time at the unscaled prior when its inertia or damping prior is wrong by an order of magnitude. In on-water field trials on the Blue Robotics BlueBoat, a twin-thruster catamaran, MOSAIC-SV transits at least $25\%$ faster and predicts its own motion with at least $56\%$ less error than its frozen engineering prior, including under an unmodeled payload. The same MOSAIC-SV system concept was also feasibly deployed on a 6.3-tonne dual outboard monohull.
\end{abstract}

\section{INTRODUCTION}
Aquatic robotic platforms such as unmanned surface vehicles (USVs) have great potential to aid in marine conservation \cite{vasilijevic_coordinated_2017}, search and rescue \cite{jorge_survey_2019}, and exploration \cite{preston_autonomous_2024}. Model-based control of a USV depends on a model of the hydrodynamic effects in its dynamics \cite{sonnenburg_modeling_2013}. Current USV controllers typically use parametric physics-based maneuvering models (e.g.\ Fossen \cite{fossen2021handbook}, MMG \cite{yasukawa_introduction_2015}) paired with predictive controllers \cite{patel_model-predictive_2023, kinjo_trajectory_2022} or learned policies \cite{meyer_taming_2020}. However, these models are non-trivial to derive and calibrate \cite{deogaonkar_data_2023}, particularly for unconventional hullforms and marine robots \cite{sutulo_development_2006}. Several approaches address this.

\begin{figure}[t]
  \centering
  \includegraphics[width=0.8\columnwidth]{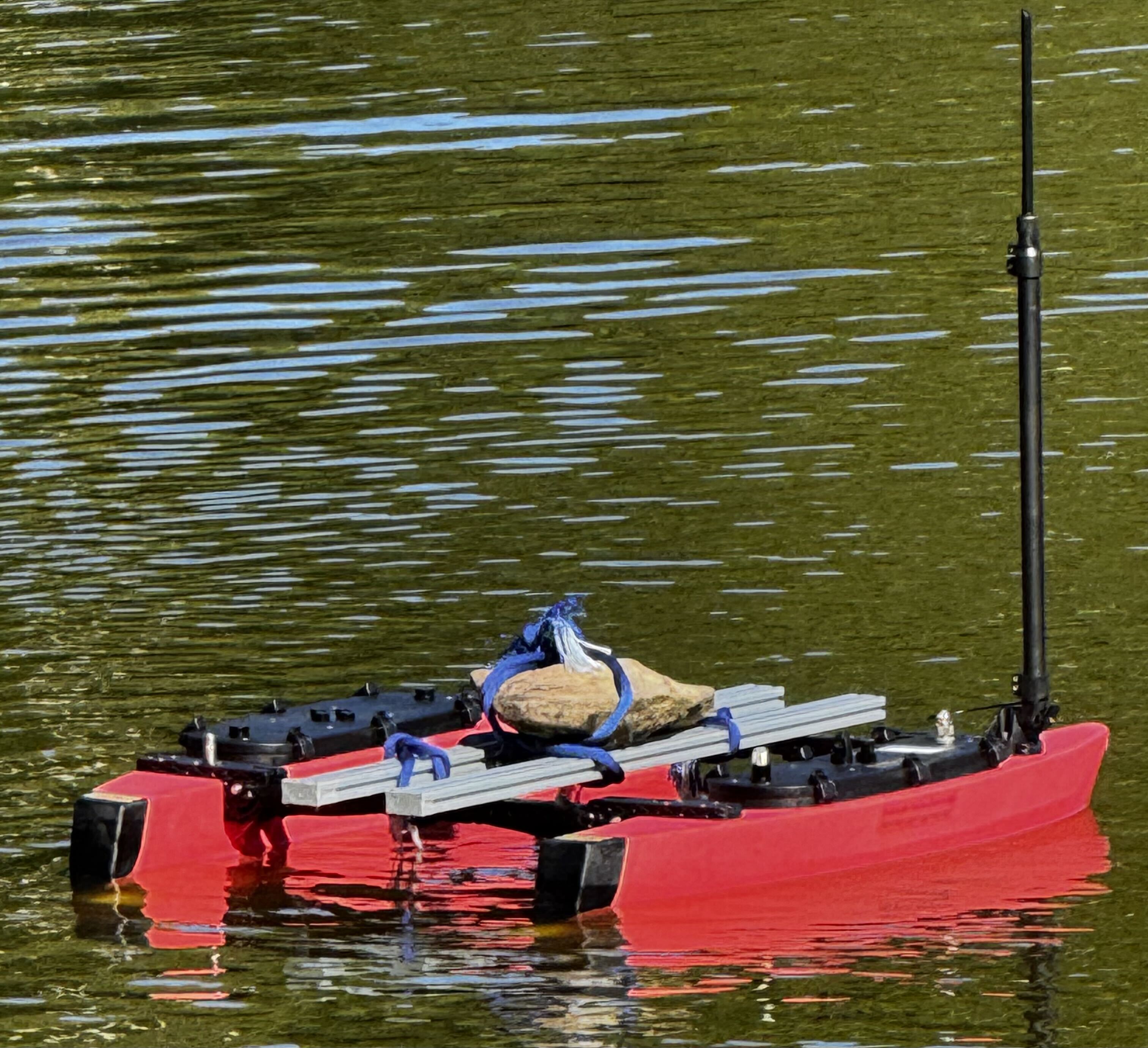}
  \caption{A Blue Robotics BlueBoat, the twin-thruster catamaran evaluated in the on-water field trials with an unmodeled extra mass (a rock).}
  \label{fig:blueboat}
\end{figure}
One approach is to identify the model \textit{a priori} from data, through parametric identification \cite{hasan_discovering_2025} or a learned dynamics model \cite{wang_learning_2024}, and give the controller the identified model. Abrougui et al.\ \cite{abrougui_modeling_2021} identify a USV model from sea trials and design heading and speed controllers on it. Wang et al.\ \cite{wang_parametric_2025} identify the maneuvering model with an extended Kalman filter on simulated excitation data before control, then run a disturbance observer alongside a nonlinear MPC in simulation. In each, the data collected and subsequent model are specific to that vessel and the conditions it was operated in; a change to the hull, payload, actuator, or sea state requires a new identification.

A second approach is to learn a correction on top of a known nominal model during operation. Zhang et al.\ \cite{zhang_experimental_2026} learn the mismatch between a nominal model and the measured yaw response online and add it to the prediction model of an MPC. Liu et al.\ \cite{liu_efficient_2026} learn per-axis residual dynamics with a sparse Gaussian process that is refined recursively inside a nonlinear MPC. Outside the marine domain, Levy et al.\ \cite{levy_meta-learning_2025} couple Kalman-filter adaptation of a dynamics model to MPPI on an off-road vehicle, but adapt along a basis meta-learned from previously collected vehicle-specific driving data. These methods, however, require a known nominal model of the test platform and require dedicated identification trials to train or calibrate the correction algorithms before deployment.

A third approach is to identify the physical model itself online, which previous work has explored as an estimation problem. Hydrodynamic coefficients are estimated as augmented states with an extended \cite{sabet_extended_2014} or unscented Kalman filter \cite{zheng_unscented_2022, meng_nonlinear_2025}, with some estimators also identifying propeller and rudder force coefficients \cite{meng_nonlinear_2025}. These studies primarily use recorded maneuvering-test data or simulation data and, when applicable, give a controller a fixed model \cite{meng_online_2024, meng_nonlinear_2025}. Additionally, these identification methods use designed maneuvers such as zig-zag, turning-circle, and spiral tests \cite{liu_efficient_2026, zheng_unscented_2022, meng_online_2024, meng_nonlinear_2025}, as every coefficient must be excited by the test, a requirement Yue et al.\ \cite{yue_online_2022} relax to interval excitation. Among the approaches, to our knowledge no published system starts a vessel from an engineering prior, identifies its physical model online during nominal operation, and plans on that model as it is re-estimated.

\subsection{Contribution}
In this work we present MOSAIC-SV (\textbf{M}aritime \textbf{O}nboard \textbf{S}ystem for \textbf{A}daptive \textbf{I}dentification and \textbf{C}ontrol of \textbf{S}urface \textbf{V}essels), a deployable real-time adaptive dynamics identification and control system. MOSAIC-SV identifies a control-sufficient dynamics model from a spec-sheet engineering prior, without dedicated identification trials, re-estimating it at every control step of a closed-loop mission while the controller plans on it. Additionally, the system is robust to poor engineering priors, remaining effective when the inertia or damping prior is wrong by an order of magnitude. MOSAIC-SV rests on two further contributions:
\begin{itemize}
  \item A physically admissible unscented Kalman filter over hydrodynamic, disturbance, and actuator parameters with a command-dependent consider projection that suppresses corrections along weak directions of the prior-scaled joint actuator and external-force map.
  \item An evaluation in simulation under static mismatch, transient changes, and prior misspecification; paired field trials on a small twin-thruster catamaran against its frozen engineering prior; and a feasibility run on a 6.3-tonne dual outboard monohull.
\end{itemize}

\section{VESSEL MODEL}
\label{sec:model}
\begin{figure}[t]
  \centering
  \begin{minipage}[b]{0.46\columnwidth}
    \centering
    \includegraphics[width=\linewidth]{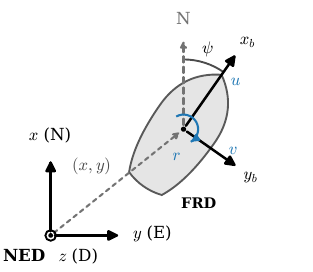}\\
  \vspace{-1em}
    {\footnotesize (a)}
  \end{minipage}\hfill
  \begin{minipage}[b]{0.52\columnwidth}
    \centering
    \includegraphics[width=\linewidth]{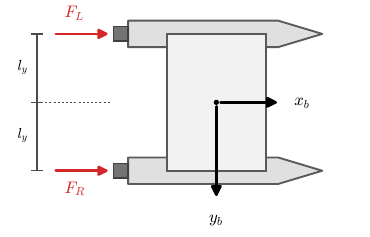}\\
  \vspace{-1em}
    {\footnotesize (b)}
  \end{minipage}
  \vspace{-.5em}
  \caption{(a) The general NED and FRD vessel frames. Heading $\psi$ is defined as clockwise from North. (b) The general form of the twin-hulled catamaran vessel class evaluated in this work, with location of forces and corresponding lever arms.}
  \label{fig:frames}
\end{figure}

We define a vessel's state $s = [\eta^\top, \nu^\top]^\top$ with standard North-East-Down (NED) global pose $\eta = [x, y, \psi]^\top$, where $x$ is North and $y$ is East, and Forward-Right-Down (FRD) body velocities $\nu = [u, v, r]^\top$ (Fig.~\ref{fig:frames}a).

\subsection{Rigid-Body Dynamics}
\label{sec:model:rb}

We define a vessel's rigid-body dynamics following a planar Fossen structure \cite{fossen2021handbook},
\begin{align}
    \dot{\eta} &= R(\psi)\,\nu \label{eq:kin}\\
    M\dot{\nu} + C(\nu)\,\nu + D(\nu)\,\nu &= \tau(a,\theta_{a}) + R^\top(\psi)\,F_{ext} \label{eq:dyn}
\end{align}
where $R(\psi)$ is the body-to-world rotation, $M = M_{RB} + M_{A}$ is the total inertia (rigid body plus added inertia), $C(\nu)$ is the Coriolis matrix derived from $M$, and $D(\nu)$ is the damping matrix, which we model with linear and diagonal quadratic components,
\begin{equation}
\label{eq:damping}
    D(\nu) = D_{lin} + \mathrm{diag}\!\left(q \odot |\nu|\right)
\end{equation}
where $\odot$ is the element-wise product. 
We define $\tau(a,\theta_{a})$ as the applied body-frame force and moment vector and $F_{ext}$ as a world-frame generalized force bias $F_{ext} = [F_x, F_y, N_z]^\top$.

\subsection{Actuator Model}
\label{sec:model:act}

In general, the actuator model has two physically distinct parts: effective force and effective yaw moment. For fixed-direction thrusters, we write the actuator model in thrust-configuration form~\cite{fossen2021handbook},
\begin{equation}
\label{eq:act}
    \tau(a,\theta_{a}) = B_a\big((\mathbf{1} + \delta\ell) \odot \ell_0\big)\;\mathrm{diag}(\mathbf{1} + \delta g)\;f_a(a)
\end{equation}
where $f_a(a) \in \mathbb{R}^{p}$ is the force each of the $p$ thrusters produces along its line of action from the actuator command $a$, $\delta g \in \mathbb{R}^{p}$ are per-thruster gain deviations from the nominal map, and $B_a \in \mathbb{R}^{3 \times p}$ resolves the forces into surge, sway, and yaw from the thruster positions and moment arms $(\mathbf{1} + \delta\ell) \odot \ell_0$. We define the deviations as the actuator parameters $\theta_a = [\delta g^\top, \delta\ell^\top]^\top$, bounded element-wise to $[-0.95,\,1]$.

The primary vessel class we evaluate in this work is a twin-hulled catamaran with two fixed thrusters (Figs.~\ref{fig:blueboat} and \ref{fig:frames}b) at lateral offset $\pm l_y$ from center with $f_a(a) = a = [F_L, F_R]^\top$. The full actuator model is
\begin{equation}
\label{eq:th_act}
    \tau(a,\theta_a) = [\,g_L F_L + g_R F_R,\; 0,\; l_y\,\ell\,(g_L F_L - g_R F_R)\,]^\top
\end{equation}
where $g_L = 1 + \delta g_L$, $g_R = 1 + \delta g_R$, and $\ell = 1 + \delta\ell$ with actuator parameters 
$\theta_a = [\delta g_L, \delta g_R, \delta\ell]^\top$.

For the dual outboard monohull tested in \S\ref{sec:results:fieldtrials}, the two outboards share one net thrust $T$ and one steering angle $\gamma$, so $a = [T, \gamma]^\top$ and (\ref{eq:act}) has a single effective thruster ($p = 1$) whose line of action is rotated by $\gamma$ and whose lever arm $l_x$ runs from the transom to the center of mass,
\begin{equation}
\label{eq:ob_act}
    \tau(a,\theta_a) = g\,[\,T\cos\gamma,\; -|T|\sin\gamma,\; l_x\,\ell\,|T|\sin\gamma\,]^\top
\end{equation}
where $g = 1 + \delta g$ and $\ell = 1 + \delta\ell$ with actuator parameters $\theta_a = [\delta g, \delta\ell]^\top$.

\section{PARAMETER ESTIMATOR DESIGN}
\label{sec:est}

We estimate the model parameters $\theta$, detailed in \S\ref{sec:est:coords}, from online state transitions during a mission, using the navigation state $s_t$ provided by the vessel's state estimator. We represent the parameter belief as a Gaussian $\mathcal{N}(\hat\theta_t, P_t)$ propagated by a scaled unscented Kalman filter (UKF), with predicted mean and covariance $\hat\theta^-_t$ and $P^-_t$, process-noise covariance $Q$, measurement $z_t$, nonlinear measurement map $h$, and observation-noise covariance $R$. We apply the standard scaled-unscented measurement update~\cite{julier_unscented_2004}, which propagates $2n+1$ sigma points of the predicted belief, where $n = \dim\theta$, to form the predicted measurement $\hat{y}$, innovation covariance $S$, and cross covariance $P_{\theta y}$. From these we compute the Kalman gain $K_0$, then update the mean $\hat\theta_t$ and covariance $P_t$ with an applied gain $K$, which is the projection of \S\ref{sec:est:sk} that derives from $K_0$,

\begin{align}
    K_0 &= P_{\theta y}\,S^{-1}, \qquad \hat\theta_t = \hat\theta^-_t + K\,(z_t - \hat{y}) \label{eq:ukf_update}\\
    P_t &= (I - KH)\,P^-_t\,(I - KH)^\top + K R_{\mathrm{eff}} K^\top \label{eq:ukf_cov}
\end{align}
where $H$ is the statistical-linear-regression matrix of $h$ \cite{lefebvre_comment_2002, garcia-fernandez_posterior_2015}, and $R_{\mathrm{eff}}$ is the observation-noise covariance $R$ (included in $S$) augmented by the regression-error covariance,
\begin{equation}
\label{eq:ukf_HReff}
    H = P_{\theta y}^\top (P^-_t)^{-1}, \qquad R_{\mathrm{eff}} = S - H P^-_t H^\top
\end{equation}

When no direction is selected, $K = K_0$. We use this Joseph covariance update (\ref{eq:ukf_cov})--(\ref{eq:ukf_HReff}) \cite{bucy_filtering_1968} because the projected gain $K$ of \S\ref{sec:est:sk} can differ from $K_0$, so the standard form of $P_t = P^-_t - K S K^\top$ can become indefinite. The Joseph covariance update guarantees $P_t \succeq 0$ for any $K$ when $P^-_t \succeq 0$ and $R_{\mathrm{eff}} \succeq 0$. If $K = K_0$, (\ref{eq:ukf_cov})--(\ref{eq:ukf_HReff}) reduces to the standard form.

\subsection{Parameter Definition}
\label{sec:est:coords}

Taking $\mathrm{vech}$ as an operation that stacks a lower-triangular matrix into a vector, we define the estimator parameters from the vessel's dynamics model as

\begin{equation}
\label{eq:theta}
\theta = \left[\,\mathrm{vech}(L_M),\; \mathrm{vech}(L_D),\; c,\; F_{ext},\; \theta_{a}\,\right]^\top \in \mathbb{R}^{21}
\end{equation}
where $L_M$ and $L_D$ are the lower-triangular Cholesky factors of $M$ and $D_{lin}$, respectively, $c \in \mathbb{R}^3$ parametrizes the quadratic-damping coefficients through $q = c \odot c$, and $F_{ext}$ and $\theta_{a}$ are the force-bias and actuator-deviation parameters. We floor the diagonal entries of both $L_M$ and $L_D$ to a small positive constant when reconstructing $M$ and $D_{lin}$, making the inertia and damping matrices positive definite by construction and ensuring that $M$ is invertible. The Cholesky factor representation therefore guarantees dissipative linear damping and avoids indefinite matrices that can arise from estimating matrix values directly. The smooth parameterization $q = c \odot c$ guarantees nonnegative quadratic-damping coefficients.

\subsection{Engineering Prior}
\label{sec:est:prior}

We construct the prior mean $\theta_0$ from the vessel's approximate specifications, not from fitted test data. The rigid-body inertia is $\mathrm{diag}(m, m, I_z)$, with $I_z = m\,(L^2 + B^2)/12$ for mass $m$, length $L$, and beam width $B$. The total inertia scales each axis by a hull-form added-mass factor, $M_0 = \mathrm{diag}\!\big(k_u\, m,\; k_v\, m,\; k_r\, I_z\big)$,

with $(k_u, k_v, k_r) = (1.10, 1.50, 1.15)$ for the catamaran and $(1.10, 1.80, 1.25)$ for the monohull. The effective damping coefficients $D_u, D_v, D_r$ are estimated drag per unit velocity at a per-axis reference velocity, with $D_v$ set as a multiple of $D_u$ for broadside damping. Each coefficient is split between the linear and quadratic terms from (\ref{eq:damping}) so that the total is reproduced at the reference velocity,

\begin{equation}
\label{eq:prior_split}
D_{lin,i} = (1 - f)\, D_i, \qquad q_i = f\, D_i / \nu_{ref,i}
\end{equation}
where $\nu_{ref} = [\,U,\; U \tan\beta,\; U/(k_{turn} L)\,]$ for a design speed $U$, a design sideslip $\beta = 10^\circ$, and a turn radius $k_{turn} = 3$ hull lengths, and $f$ is the quadratic share of the damping at the reference velocity, set from the hull's Froude regime ($f = 0.70$ for the catamaran, $0.85$ for the monohull). The $F_{ext}$ and $\theta_a$ priors are zero.

We set the prior covariance as $P_0 = \mathrm{diag}(\sigma)^2$, with the prior standard deviation $\sigma_i$ of each parameter set to $20\%$ of $\mathrm{vech}(L_M)$, $35\%$ of $\mathrm{vech}(L_D)$, $5\%$ of $c$, $0.1$ on each $\theta_{a}$, and a yaw moment equal to $5\%$ of the vessel's maximum yaw moment. The surge and sway $F_{ext}$ standard deviations are set to $25\%$ of the vessel's max thrust. This ordering lets an innovation be absorbed by the hull parameters before the actuator deviations or the yaw bias, while surge and sway forces remain free to track external disturbances. Every standard deviation is floored at a small positive value, including the zero-mean off-diagonal Cholesky parameters, so that $P_0 \succ 0$ and $P_0^{-1/2}$ in (\ref{eq:consider}) is defined.

\subsection{Process Model}
\label{sec:est:ou}

We define the process model as a discrete Ornstein-Uhlenbeck prediction \cite{sarkka_applied_2019} with a blockwise rate vector $\lambda$,
\begin{align}
\hat\theta^-_{t} &= \hat\theta_{t-1} + \Lambda\,(\theta_0 - \hat\theta_{t-1}) \label{eq:ou}\\
P^-_{t} &= (I - \Lambda)\,P_{t-1}\,(I - \Lambda) + Q \label{eq:ou_cov}
\end{align}

where $\Lambda = \mathrm{diag}(\lambda)$, and we choose $Q$ as a diagonal process-noise covariance. For $F_{ext}$ parameters, we select $0 < \lambda_i < 1$ and set $Q_{ii} = [1 - (1-\lambda_i)^2]\,P_{0,ii}$, which holds the prior variance stationary at $P_{0,ii}$ under
prediction alone. These parameters revert toward their priors with time constant $-\Delta t/\log(1-\lambda_i) \approx \Delta t/\lambda_i$. For inertia, linear damping, quadratic damping, and $\theta_a$, we set $\lambda_i = 0$ and choose positive process-noise variances, giving random walks without reversion. 

This process model allows the external-force bias to track changing disturbances while retaining accumulated hydrodynamic and actuator corrections.

\subsection{Measurement Update and Consider Projection}
\label{sec:est:sk}

We define $\Phi_{\Delta t}(s, a;\,\theta)$ as one fourth-order Runge-Kutta step of (\ref{eq:kin})--(\ref{eq:dyn}) over the interval $\Delta t$ under a zero-order hold command. We define the measurement model $h$ as the one-step prediction of the current navigation state from the preceding one,
\begin{align}
\label{eq:meas}
h(\theta) &= \Phi_{\Delta t}(s_{t-1}, a_{t-1};\,\theta), \nonumber\\
z_t &= s_t = h(\theta) + w_t, \qquad w_t \sim \mathcal{N}(0, R)
\end{align}
where $s_{t-1}$ and $s_t$ are known from the vessel's state estimator and the observation-noise covariance $R$ is diagonal. We construct $R$ from the state estimator's reported uncertainty when available, static values otherwise, and inflate it by a fixed factor to account for time-correlated state estimator errors across transitions. We calculate the heading component of the innovation $z_t - \hat{y}$ as the principal angular difference.
We regularize the innovation covariance $S$ before inversion as $S_\rho = S + \rho\,\|S\|_F\,I$, with $\rho$ proportional to machine precision, to improve numerical conditioning.

The plain Kalman gain $K_0 = P_{\theta y} S_\rho^{-1}$, (\ref{eq:ukf_update}) with $S_\rho$ in place of $S$, allocates the innovation among parameters according to the predicted covariance and measurement sensitivity. However, when two parameters produce the same force, the measurement cannot reliably distinguish them, and the gain splits the correction between them according to the predicted covariance. We address this by suppressing corrections along force-parameter $[F_{ext}^\top, \theta_a^\top]^\top$ directions with weak instantaneous sensitivity relative to their prior standard deviations. We determine these directions from the force-map Jacobian scaled by the prior standard deviation of each parameter, for the catamaran,
\begin{equation}
\label{eq:jforce}
J_t = A\,\big[\,R^\top(\psi_{t-1}),\; J_{a,t}\,\big] \in \mathbb{R}^{3 \times 6}, \quad J_w = J_t\,P_{0,f}^{1/2}
\end{equation}
where $A = \mathrm{diag}(1, 1, 1/l_y)$ converts every row into force units, $R^\top(\psi_{t-1})$ is the sensitivity to $F_{ext}$ at heading $\psi_{t-1}$, and $J_{a,t}$ is the Jacobian of (\ref{eq:act}) with respect to $\theta_{a}$ at the command $a_{t-1}$ and predicted mean $\hat\theta^-_t$. $J_w$ is the force-map Jacobian scaled by $P_{0,f}^{1/2}$, the block of $P_0^{1/2}$ on the force parameters $[F_{ext}^\top, \theta_a^\top]^\top$, and measures sensitivity relative to each force-parameter's prior standard deviation.

We define $V_0$ as the matrix of all right singular vectors of $J_w$ whose singular values are zero or below a fixed fraction of the largest. With zero rows for the inertial and damping parameters in $V_0$, the projected Kalman gain is
\begin{equation}
\label{eq:consider}
K = (I - \Pi)\,K_0, \qquad \Pi = P_0^{1/2}\,V_0 V_0^\top\,P_0^{-1/2}
\end{equation}
following the Schmidt-Kalman consider filter \cite{schmidt_application_1966} and its partial-update forms \cite{brink_partial-update_2017, brink_unscented_2018}. Observability-informed variants \cite{ramos_observability_2021} adapt how strongly selected states are corrected; our rule instead selects force-parameter directions per transition from the prior-scaled instantaneous force map. The projection removes the mean correction along the selected directions, leaves the inertial and damping parameter rows of $K_0$ unchanged, and keeps the uncertainty and cross-covariances of the selected directions in $P_t$, where they continue to propagate.

We evaluate (\ref{eq:ukf_cov}) from paired sigma points using the applied gain $K$. We generate the sigma points from $P^-_\varepsilon = P^-_t + \varepsilon I$, where $\varepsilon$ is a small positive constant that ensures $P^-_\varepsilon$ is positive definite. We use $P^-_\varepsilon$ and $S_\rho$ in place of $P^-_t$ and $S$ throughout (\ref{eq:ukf_cov})--(\ref{eq:ukf_HReff}). This construction is positive semidefinite for any gain, including the projected $K$.

\section{CONTROLLER DESIGN}
\label{sec:ctrl}

We plan vessel commands with model predictive path integral control (MPPI)~\cite{williams_aggressive_2016} on the transition $\Phi_{\Delta t}$ defined in (\ref{eq:meas}).
At each control step we define a nominal control sequence $\mathcal{U} = [a_0, \ldots, a_{N-2}]$ over a horizon of $N$ states and sample $N_s$ perturbed sequences,
\begin{equation}
\label{eq:mppi_sample}
\mathcal{U}^k = \mathcal{U} + E^k, \qquad E^k_i \sim \mathcal{N}(0, \Sigma),
\quad k = 1, \ldots, N_s
\end{equation}

where $E^k = [E^k_0, \ldots, E^k_{N-2}]$ is the perturbation sequence and $\Sigma$ is a diagonal per-command sampling covariance. We roll each sequence out from the current navigation state $s_t$ using the current estimated dynamics model $\hat\theta_t$ from \S\ref{sec:est}, which we assume is fixed over the horizon. We do not propagate its covariance $P_t$ through the rollouts or into the objective. We clip actuator commands element-wise to the actuator limits, $\tilde{a}^k_i = \mathrm{clip}(a_i + E^k_i) \in [a_{\min}, a_{\max}]$, and roll out $s^k_0 = s_t$, $s^k_{i+1} = \Phi_{\Delta t}\big(s^k_i,\; \tilde{a}^k_i;\; \hat\theta_t\big)$ for $i = 0, \ldots, N-2$. We score each rollout with a cost $J_k$ that penalizes quadratic goal-state error along the horizon and at its end, control effort and step-to-step command changes, misalignment between heading and goal bearing scaled by distance to the goal, and proximity to obstacles. We update the nominal sequence with the softmax-weighted average of the perturbations,
\begin{align}
\label{eq:mppi_update}
w_k &= \frac{\exp\!\big(-(J_k - J_{\min})/\lambda_{\mathrm{eff}}\big)}
           {\sum_j \exp\!\big(-(J_j - J_{\min})/\lambda_{\mathrm{eff}}\big)}, \nonumber\\
\mathcal{U} &\leftarrow \mathrm{clip}\Big(\mathcal{U} + \textstyle\sum_k w_k E^k\Big)
\end{align}
where $J_{\min} = \min_k J_k$ and $\lambda_{\mathrm{eff}}$ is the temperature defined in \S\ref{sec:ctrl:temp}. We clip the updated sequence $\mathcal{U}$ separately from the rollouts because the update uses the unclipped perturbations $E^k$ and can exceed the actuator limits. We apply the first command of the updated sequence and shift the sequence by one step to warm start the next control step.

\subsection{Adaptive Temperature}
\label{sec:ctrl:temp}

With a fixed temperature, the weights of (\ref{eq:mppi_update}) depend on the scale of the sampled cost differences, which varies over a mission. Carius et al.~\cite{carius_constrained_2022} normalize the costs by the batch range; we instead scale the temperature by the median-to-minimum spread, subject to a positive floor,

\begin{equation}
\label{eq:lam}
\lambda_{\mathrm{eff}} = \max\!\left(\lambda_{\mathrm{floor}},\;
\lambda_{\mathrm{rel}}\,\big(\mathrm{median}(J) - \min(J)\big)\right)
\end{equation}
where the median and minimum are taken over the sampled costs $\{J_k\}_{k=1}^{N_s}$, $\lambda_{\mathrm{rel}} > 0$ sets the temperature as a fraction of the cost spread, and $\lambda_{\mathrm{floor}} > 0$ bounds it away from zero. While the floor is inactive, the weights are invariant to a common offset or positive scaling of the costs. The median limits the influence of a minority of high-cost obstacle rollouts on the temperature.

\section{RESULTS}
\label{sec:results}
We evaluate the estimator-controller system from \S\ref{sec:est}--\S\ref{sec:ctrl}, MOSAIC-SV, in closed-loop simulation against two non-adaptive planning models, and on water against the frozen engineering prior.

\subsection{Simulation Setup}
\label{sec:results:setup}

In simulation, we use a calibrated model of CyberShip~II, a 1:70-scale supply-vessel model whose inertia, Coriolis, and damping coefficients are published from tank identification~\cite{skjetne_modeling_2004, hasan_discovering_2025}. We replace its fully actuated thrust allocation with the differential twin-thruster actuation of (\ref{eq:th_act}), with per-thruster limits sized to the hull at a cruise speed of 1.5~m/s. 

\begin{figure}[t]
  \centering
  \includegraphics[width=\columnwidth]{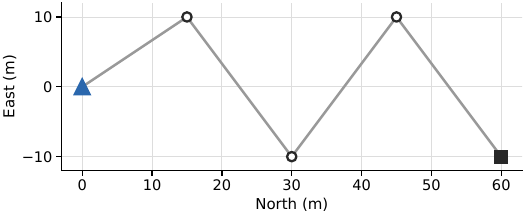}
  \vspace{-2em}
  \caption{Simulation course in the world frame. The triangle marks the start, open circles waypoints passed under way, and the square the stop waypoint.}
  \label{fig:missions}
\end{figure}

We set the simulation's plant as the calibrated model, and perturb its parameters to generate mismatch and transients for the experiments. In the experiments, we compare three different planning models: MOSAIC-SV, the calibrated model, and the \emph{frozen} engineering prior from \S\ref{sec:est:prior} integrated with the MPPI controller from \S\ref{sec:ctrl}. Because the calibrated and frozen models lack adaptation, $\hat \theta_t$ is the same at every control step.

The simulated experiments all run the course shown in Fig.~\ref{fig:missions} under one of three families of perturbation, shown in Table~\ref{tab:families}. \emph{Static mismatch} runs different perturbed plants as the simulated plant. \emph{Transient effects} change inertia, damping, disturbance force, or delivered thrust on the port, starboard, or both thrusters part-way through the run, using step or ramp profiles. \emph{Prior sensitivity} keeps the plant unperturbed and scales the engineering prior MOSAIC-SV starts from, on inertia, damping, or both. Each variation within a family is run over twelve trials, with each trial having a different pseudo-random number generator (PRNG) seed. The twelve PRNG seeds are consistent between every experiment variation run.

\begin{table}[t]
\centering
\vspace{0.5em}
\caption{Perturbation families}
\vspace{-1em}
\label{tab:families}
\setlength{\tabcolsep}{1.0pt}
\scriptsize
\begin{tabular}{@{}p{.47in}p{.48in}p{1.03in}p{1.30in}@{}}
\toprule
\raggedright Family & \raggedright Axis & \raggedright Perturbation & \raggedright Levels \tabularnewline
\midrule
\raggedright Static & \raggedright mass & \raggedright Inertia $\times$ level & \raggedright $2^{-3},\ldots,2^{4}$ \tabularnewline
\raggedright Static & \raggedright cross & \raggedright Inertia $\times$, damping $\div$ & \raggedright $2^{-3},\ldots,2^{4}$ \tabularnewline
\raggedright Static & \raggedright damping & \raggedright Damping $\times$ level & \raggedright $2^{-3},\ldots,2^{4}$ \tabularnewline
\raggedright Static & \raggedright combined & \raggedright Inertia/damping $\times$; 45$^\circ$ & \raggedright $2^{-3},\ldots,2^{4}$; force 0--3.2\,N$^{\ddagger}$ \tabularnewline
\raggedright Static & \raggedright current & \raggedright Force at 45/0/90$^\circ$ & \raggedright 0--3.2\,N, step 0.4\,N \tabularnewline
\midrule
\raggedright Transient & \raggedright mass & \raggedright Inertia $\times$ level & \raggedright $1/4$, $1/2$, $1.5$, $2$, $3$, $4$, $6$, $8$ \tabularnewline
\raggedright Transient & \raggedright damping & \raggedright Damping $\times$ level & \raggedright $1/8$, $1/4$, $1/2$, $2$, $4$, $8$ \tabularnewline
\raggedright Transient & \raggedright current & \raggedright Force at 45$^\circ$ & \raggedright 0.4, 0.8, 1.2, 2, 2.8\,N \tabularnewline
\raggedright Transient & \raggedright port & \raggedright Thrust fraction, port & \raggedright $1/4$, $1/2$, $0.8$, $0.9$, $1^\dagger$ \tabularnewline
\raggedright Transient & \raggedright stbd & \raggedright Thrust fraction, stbd & \raggedright $1/4$, $1/2$, $0.8$, $0.9$, $1^\dagger$ \tabularnewline
\raggedright Transient & \raggedright both & \raggedright Thrust fraction, both & \raggedright $1/4$, $1/2$, $5/8$, $3/4$, $0.9$, $1^\dagger$ \tabularnewline
\midrule
\raggedright Prior & \raggedright inertia & \raggedright Prior inertia scale & \raggedright $2^{-3},\ldots,2^{4}$ \tabularnewline
\raggedright Prior & \raggedright damping & \raggedright Prior damping scale & \raggedright $2^{-3},\ldots,2^{4}$ \tabularnewline
\raggedright Prior & \raggedright both & \raggedright Both prior scales & \raggedright $2^{-3},\ldots,2^{4}$ \tabularnewline
\bottomrule
\end{tabular}
\par\vspace{4pt}\noindent
\begin{minipage}{\columnwidth}\scriptsize\raggedright
Static conditions are applied before the run; transients start at 34.8 s as a step or as a ramp over 41.8 s. Static and transient families compare MOSAIC-SV with the calibrated model and the frozen prior. The prior family runs MOSAIC-SV only. We run twelve seeds per level with a 400 s run limit and a 2 m goal radius. $^\dagger$ indicates a level that was only executed as a step change. $^\ddagger$ Combined levels $2^{-3}\ldots2^{4}$ pair with 45$^\circ$ forces 0, 0.4, 0.8, 1.2, 1.6, 2.0, 2.4, 3.2\,N.
\end{minipage}\vspace{2pt}
\end{table}

\begin{figure*}[t]
  \centering
  \includegraphics[width=\textwidth]{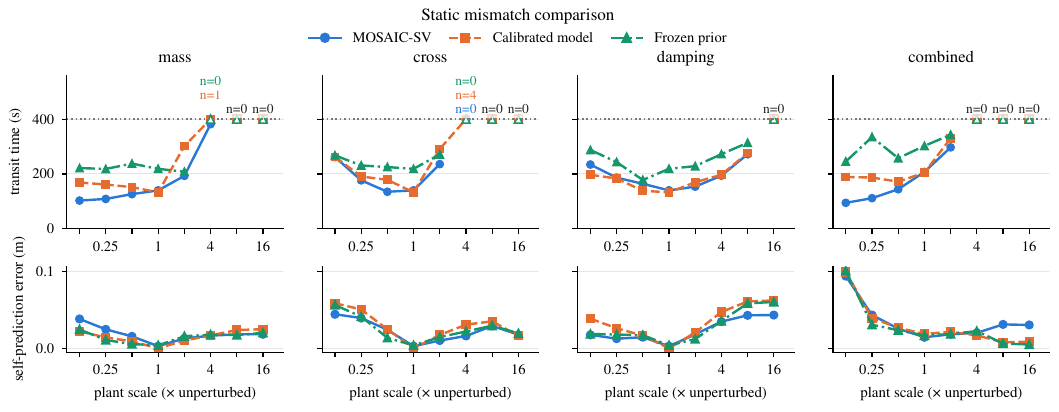}
  \vspace{-2em}
  \caption{Static mismatch. Top: median transit time per plant level. $n$ gives the number of seeds that completed the run when there were incomplete runs. Bottom: median self-prediction error. Levels are the scale applied to the unperturbed plant.}
  \label{fig:static}
\end{figure*}

We report three metrics. \emph{Completion} is the vessel landing within 2~m of the final waypoint before the run time limit, reported as $\Delta c$, the difference in non-completion rate between MOSAIC-SV and each baseline in percentage points. \emph{Transit time} is the time the vessel took to complete a run; a run that does not complete is assigned the run time limit as its transit time. This is reported as $\Delta_{\mathrm{tr}}$, the percent difference between the two models' median transit times.  \emph{Self-prediction error} scores each planning model on how well it predicted its future positions. For each run, we record the state and command applied each timestep. From each state $s_t$ we roll the planning model forward $N_{sp}$ steps under the commands that were applied, $\hat s_{t+j+1} = \Phi_{\Delta t}\big(\hat s_{t+j},\, a_{t+j};\, \hat\theta_t\big),~\hat s_{t} = s_t,$

for $j = 0,\ldots,N_{sp}-1$, and take the position error $e_t = \lVert \hat p_{t+N_{sp}} - p_{t+N_{sp}} \rVert$, where $\hat p_{t+N_{sp}}$ is the rolled-out final position and $p_{t+N_{sp}}$ is the actual final planar position $(x, y)$ of that horizon. We take the median of $e_t$ over the run with $N_{sp}\Delta t = 1$~s, evaluated every $0.5$~s, and report $\Delta e$, the median seed-paired difference in this error as a percentage of the baseline's median. For MOSAIC-SV, $\hat\theta_t$ is the estimated dynamics model; for the calibrated model and the frozen prior $\hat\theta_t$ is constant.

\subsection{Simulation Results}
\subsubsection{Static Mismatch}
\begin{table}[t]
\centering
\vspace{0.5em}
\caption{Static simulation comparison}
\vspace{-1em}
\label{tab:pooled-sim}
\setlength{\tabcolsep}{4.5pt}
\footnotesize
\begin{tabular}{@{}lrrrrrrr@{}}
\toprule
& & \multicolumn{3}{c}{vs. calibrated} & \multicolumn{3}{c}{vs. frozen} \\
\cmidrule(lr){3-5}\cmidrule(l){6-8}
Condition & $n$ & $\Delta_{\mathrm{tr}}$\% & $\Delta c$ pp & $\Delta e$\% & $\Delta_{\mathrm{tr}}$\% & $\Delta c$ pp & $\Delta e$\% \\
\midrule
mass & 96 & -25.2 & -11.5 & +17.3 & -27.5 & -12.5 & +3.1 \\
cross & 96 & -0.6 & +4.2 & -29.0 & -2.7 & +0.0 & -8.9 \\
damping & 96 & +2.1 & +0.0 & -41.3 & -26.2 & +0.0 & -8.5 \\
combined & 96 & -5.9 & +0.0 & +12.8 & -26.0 & +0.0 & +11.7 \\
\textbf{hull, all} & 384 & -1.5 & -1.8 & -7.1 & -26.0 & -3.1 & -5.8 \\
\midrule
current 45$^\circ$ & 108 & +0.7 & +0.0 & -23.2 & -28.0 & -22.2 & -11.0 \\
current 0$^\circ$ & 108 & +5.5 & +0.0 & -39.8 & -32.6 & -44.4 & -38.5 \\
current 90$^\circ$ & 108 & +4.6 & +0.0 & -17.9 & -8.7 & -22.2 & -10.9 \\
\textbf{current, all} & 324 & +5.5 & +0.0 & -23.2 & -22.5 & -29.6 & -12.7 \\
\midrule
\textbf{static, all} & 708 & -0.6 & -1.0 & -15.4 & -23.2 & -15.3 & -9.7 \\
\bottomrule
\end{tabular}
\par\vspace{4pt}\noindent
\begin{minipage}{\columnwidth}\scriptsize\raggedright
$\Delta_{\mathrm{tr}}$: median transit time percent difference; $\Delta c$: non-completion-rate percentage-point difference; $\Delta e$: self-prediction error percent difference, paired by seed. Negative values favor MOSAIC-SV. $n$ = episodes per model. The \textbf{all} rows pool all trials within the condition group.
\end{minipage}\vspace{2pt}
\end{table}

Under static mismatch (Table~\ref{tab:pooled-sim}, Fig.~\ref{fig:static}), MOSAIC-SV successfully recovers the transit performance of the calibrated model. Pooled over all static conditions, its median transit time is within $0.6\%$ of the calibrated model's in either direction, with a $1.0$~pp lower non-completion rate and $15.4\%$ lower self-prediction error. The largest gain is under mass mismatch, where MOSAIC-SV transits $25.2\%$ faster; the largest deficit is under external current, where it transits $5.5\%$ slower while predicting $23.2\%$ better. Compared to the frozen engineering prior, MOSAIC-SV transits faster in every static condition, $23.2\%$ faster pooled, with $15.3$~pp fewer non-completions and $9.7\%$ lower self-prediction error.

\subsubsection{Transient Effects}
\begin{table}[t]
\centering
\caption{Transient simulation comparison}
\vspace{-1em}
\label{tab:pooled-transients}
\setlength{\tabcolsep}{4.5pt}
\footnotesize
\begin{tabular}{@{}lrrrrrrr@{}}
\toprule
& & \multicolumn{3}{c}{vs. calibrated} & \multicolumn{3}{c}{vs. frozen} \\
\cmidrule(lr){3-5}\cmidrule(l){6-8}
Condition & $n$ & $\Delta_{\mathrm{tr}}$\% & $\Delta c$ pp & $\Delta e$\% & $\Delta_{\mathrm{tr}}$\% & $\Delta c$ pp & $\Delta e$\% \\
\midrule
Step mass & 96 & -11.8 & +25.0 & -24.2 & +5.0 & +12.5 & -26.7 \\
Step damping & 72 & +9.9 & +0.0 & -38.9 & -31.1 & +0.0 & -18.2 \\
Step current & 60 & +8.6 & +0.0 & -31.7 & -30.1 & -20.0 & -16.1 \\
Step port & 60 & +3.4 & +0.0 & -23.5 & -29.4 & +0.0 & -37.8 \\
Step stbd & 60 & +1.5 & +0.0 & -4.7 & -35.5 & +0.0 & -22.9 \\
Step sym. & 72 & +6.7 & +0.0 & +23.1 & -30.4 & +0.0 & -5.6 \\
\textbf{Step all} & 420 & +8.0 & +5.7 & -19.1 & -23.1 & +0.0 & -17.9 \\
\midrule
Ramp mass & 96 & +2.8 & +0.0 & -42.7 & +14.2 & -8.3 & -47.8 \\
Ramp damping & 72 & +3.7 & +0.0 & -19.3 & -29.2 & +0.0 & -12.8 \\
Ramp current & 60 & +3.6 & +0.0 & -31.9 & -19.0 & -20.0 & -20.8 \\
Ramp port & 48 & -3.8 & +0.0 & -1.6 & -30.0 & +0.0 & -30.2 \\
Ramp stbd & 48 & +5.1 & +0.0 & +9.8 & -31.2 & +0.0 & -11.1 \\
Ramp sym. & 60 & +3.0 & +0.0 & +38.1 & -29.5 & +0.0 & -22.4 \\
\textbf{Ramp all} & 384 & -1.4 & +0.0 & -7.3 & -26.7 & -5.2 & -21.4 \\
\midrule
\textbf{All transients} & 804 & +2.1 & +3.0 & -16.6 & -25.0 & -2.5 & -19.6 \\
\bottomrule
\end{tabular}
\par\vspace{4pt}\noindent
\begin{minipage}{\columnwidth}\scriptsize\raggedright
$\Delta_{\mathrm{tr}}$: median transit time percent difference; $\Delta c$: non-completion-rate percentage-point difference; $\Delta e$: self-prediction error percent difference, paired by seed. Negative values favor MOSAIC-SV. $n$ = episodes per model. The \textbf{all} rows pool all trials within the condition group.
\end{minipage}\vspace{2pt}
\end{table}

Under transient effects (Table~\ref{tab:pooled-transients}), MOSAIC-SV also recovers the calibrated model's transit performance. Pooled, its median transit time is within $2.1\%$ (within $8.0\%$ under step and $1.4\%$ under ramp effects), it is never more than $9.9\%$ slower in any single condition, and it predicts its future positions with $16.6\%$ less error. The exception is completion under step mass changes, where MOSAIC-SV failed to complete $25$~pp more runs than the calibrated model. Compared to the frozen prior, MOSAIC-SV transits $25.0\%$ faster pooled and predicts with $19.6\%$ less error; the only conditions where it transits slower than the frozen prior are the step and ramp mass changes ($5.0\%$ and $14.2\%$).

\subsubsection{Prior Sensitivity}
\begin{figure}[t]
  \centering
  \includegraphics[width=\columnwidth]{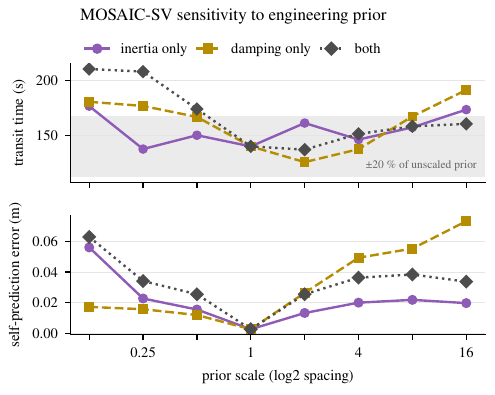}
  \vspace{-2em}
  \caption{MOSAIC-SV sensitivity to the engineering prior. Evaluated using an unperturbed plant and scaling the prior inertia, damping, or both. The band marks $\pm 20\%$ of the transit time at the unscaled prior.}
  \label{fig:prior}
\end{figure}
Prior sensitivity (Fig.~\ref{fig:prior}) measures the effect of deviating from the initial unscaled prior; the flatter the curve, the less sensitive MOSAIC-SV is to a deviation in that prior direction. As shown, MOSAIC-SV is robust to large-scale misspecification of the engineering prior. MOSAIC-SV can have a deviation between $\times 0.25$ and $\times 8$ in inertia, $\times 0.5$ and $\times 8$ in damping, as well as $\times 1$ and $\times 16$ in both inertia and damping, and still stay within $20\%$ of its transit performance at the unscaled prior. Additionally, MOSAIC-SV can have a deviation between $\times 0.25$ and $\times 16$ in inertia, $\times 0.125$ and $\times 2$ in damping, and $\times 0.25$ and $\times 16$ in both inertia and damping, and maintain a self-prediction error below 0.04~m at every scale in those ranges. 

\subsection{Field Trials}
\label{sec:results:fieldtrials}
\begin{figure*}[t]
  \centering
  \includegraphics[width=\textwidth]{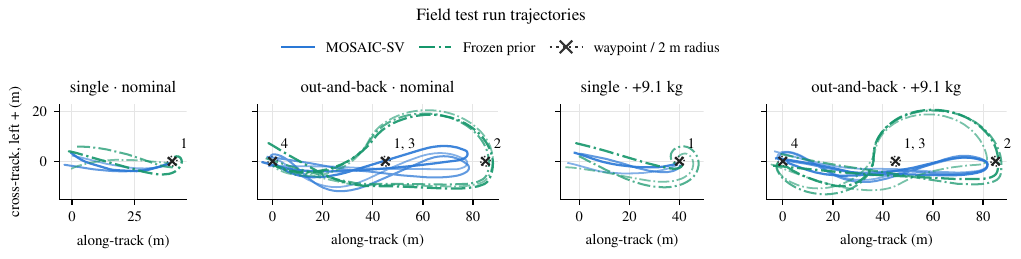}
  \vspace{-2em}
  \caption{Field test run trajectories. Crosses mark waypoints with their 2~m arrival radius.}
  \label{fig:field}
\end{figure*}

The primary field platform is a Blue Robotics BlueBoat, a 19.5~kg, 1.2~m long, twin-hull catamaran with one fixed thruster per hull (Fig.~\ref{fig:blueboat}).
We conducted field tests running either MOSAIC-SV or its frozen engineering prior (no calibrated model was available). The first mission was a single-waypoint mission that had the boat travel to a waypoint 40~m away and stop. The second was an out-and-back mission that had the boat travel to two waypoints in a line, one 45~m and one 85~m away, then loop back to the start through the same waypoints. We ran three of each mission both unloaded and with a 9.1~kg mass attached to the vessel that was not reflected in the models (a $\times 1.47$ mass mismatch). We measured completion, transit time, and self-prediction error in the same manner as the simulation experiments. 

Both MOSAIC-SV and the frozen engineering prior completed all of their runs. 

MOSAIC-SV universally transited faster and with lower self-prediction error (Table~\ref{tab:field-pairs-pct}), maintaining at least $25.3\%$ faster transit times and at least $56.9\%$ lower self-prediction error.
Additionally, Fig.~\ref{fig:field} shows that MOSAIC-SV produced tighter trajectories for all runs.

\begin{table}[t]
\centering
\caption{Field trial comparison}
\vspace{-1em}
\label{tab:field-pairs-pct}
\setlength{\tabcolsep}{4.5pt}
\footnotesize
\begin{tabular}{@{}lrrr@{}}
\toprule
& \multicolumn{3}{c}{vs. frozen} \\
\cmidrule(lr){2-4}
Cell & $n$ & $\Delta_{\mathrm{tr}}$\% & $\Delta e$\% \\
\midrule
Single nom. & 3 & -40.3 & -59.9 \\
O/B nom. & 3 & -40.0 & -69.7 \\
Single +9.1 kg & 3 & -42.9 & -56.9 \\
O/B +9.1 kg & 3 & -25.3 & -70.3 \\
\bottomrule
\end{tabular}
\par\vspace{4pt}\noindent
\begin{minipage}{\columnwidth}\scriptsize\raggedright
$\Delta_{\mathrm{tr}}$: median transit time percent difference; $\Delta e$: self-prediction error percent difference, paired by run. Negative values favor MOSAIC-SV. $n$ = pairs per cell.
\end{minipage}\vspace{2pt}
\end{table}

\begin{figure}[t]
  \centering
  \includegraphics[width=\columnwidth]{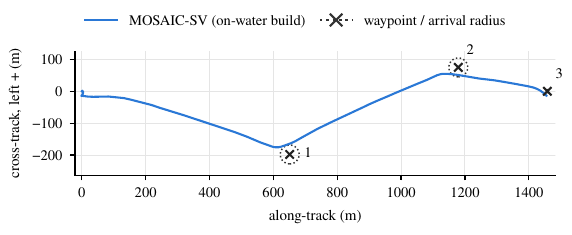}\\
  {\footnotesize (a)}\\[2pt]
  \includegraphics[width=\columnwidth]{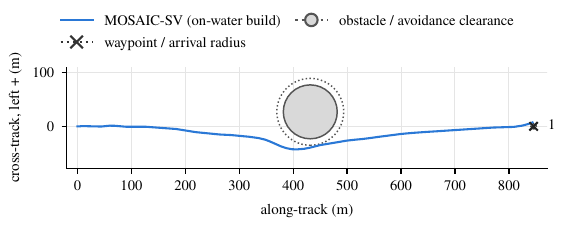}\\
  {\footnotesize (b)}
  \vspace{-1em}
  \caption{On-water runs of the 6.3-tonne dual outboard monohull. (a) Three-waypoint run. (b) Single-obstacle run. Crosses mark waypoints with their arrival radius; the circle marks the obstacle with its avoidance clearance.}
  \label{fig:dualtrial}
\end{figure}

We also ran MOSAIC-SV on a 6.3-tonne, 12~m long, dual outboard monohull on a 1.6~km three-waypoint course and a 0.85~km obstacle avoidance course at planing speed (approximately 35 knots) (Fig.~\ref{fig:dualtrial}). Its prior was produced from \S\ref{sec:est:prior} with the hull's length, beam, mass, and rated engine power, without additional offline hydrodynamic identification. This run establishes that our approach can feasibly deploy across a more than $300\times$ range of vessel mass and onto a steerable-outboard actuator model. Due to limitations with vessel availability, comparisons to other models were not conducted on this vessel. 

\section{DISCUSSION}
\subsection{Self-Prediction and Transit Time}
As shown in Table~\ref{tab:pooled-sim}, self-prediction error and transit time do not necessarily correlate. For example, under mass mismatch MOSAIC-SV transits $25.2\%$ faster than the calibrated model while having $17.3\%$ higher self-prediction error. Alternatively, under external current it predicts $23.2\%$ better while transiting $5.5\%$ slower. This is because the two metrics answer different questions: consistency of the belief vs. quality of the resulting plan. Self-prediction measures whether the model's one-second predictions match what the vessel did, while transit time measures whether the commands those rollouts executed were good over the mission. A small, stable error in the model minimally impacts transit time with a controller that re-plans every step; alternatively a slow model may track well but result in a conservative plan. 

\subsection{Trends Across Perturbations}
MOSAIC-SV performs best when the mismatch lies in directly excited parameters with a wide prior (Table~\ref{tab:pooled-sim}). For example, inertia is excited by every acceleration along its axis and carries a wide prior (\S\ref{sec:est:prior}), and mass mismatch is where MOSAIC-SV gains most over the calibrated model (Table~\ref{tab:pooled-sim}). 

Alternatively, the current is represented as the force bias $F_{ext}$, which \S\ref{sec:est:ou} treats as a fast state reverting toward zero, and under external current MOSAIC-SV transits $5.5\%$ slower than the calibrated model while predicting $23.2\%$ better. A fast bias is generally beneficial, as it absorbs whatever the hydrodynamic blocks cannot explain and lets the controller act on it quickly; however, reversion also means a persistent current is partly forgotten and re-learned throughout the run.

\subsection{Limitations}
\subsubsection{Sim-to-Real Gap}
Two effects were observed on water that are not present in the simulated plant. First, the vessel's yaw response was weaker than its engineering model predicts. The differential thrust produced less turning moment than the lever-arm geometry implies, which was absorbed into the estimator as yaw damping and lever-arm parameters. Second, the thrusters have a start-up dead time and a low-command dead band, which are poorly explained by any constant-gain model and which the estimator attributes to a phantom force. Neither effect is present in the simulation runs: the simulated plant produces exactly the yaw moment its lever-arm geometry implies, and its thrusters have no start-up delay or dead band.

\subsubsection{Innovation Aliasing}
An innovation is the mismatch between the predicted and observed transition, and several parameters can explain the same mismatch. For example, within the force parameters, a thrust shortfall and an opposing surge force produce the same surge innovation. The consider projection of \S\ref{sec:est:sk} withholds the correction when the current command cannot separate them, but is evaluated per interval and not over the entire run. Therefore the estimated parameters are control-sufficient but are not literal measurements of the vessel hydrodynamic parameters. This is also why the out-of-model sim-to-real effects land in their respective blocks.

\section{CONCLUSIONS}
We present MOSAIC-SV, a deployable real-time adaptive control system that can identify a control-sufficient dynamics model, without dedicated identification trials, and plan on it. 

In simulation, the system recovers the transit performance of a calibrated model under static mismatch or transient changes and tolerates order-of-magnitude scaling of its inertia and damping priors. On water, MOSAIC-SV transited faster and predicted its own motion better than a frozen prior in every paired run, including under an unmodeled payload. The same system concept was also feasibly deployed to control a 6.3-tonne vessel.

Future work will extend the trials to further hull and actuator classes, propagate the parameter covariance into the MPPI rollouts so the controller can hedge along unresolved directions, and extend the formulation from 3-DoF to 6-DoF to account for vessel stability.

\section{ACKNOWLEDGMENTS}
This work contains simulation, validation, and figure-plotting scripts generated by and debugged using assistance from agentic AI coding tools Claude and ChatGPT. We would like to acknowledge Moraine State Park – Pennsylvania Department of Conservation and Natural Resources for their assistance in providing a testing space for our experiments. This work is Patent Pending.

\bibliographystyle{IEEEtran}
\bibliography{references/references-fixed}

\end{document}